\documentclass[prd,amsmath,amssymb]{revtex4}

\usepackage{graphicx}
\usepackage{dcolumn}
\usepackage{bm}
\usepackage{cases}
\usepackage{mathtools}
\DeclarePairedDelimiter\bra{\langle}{\rvert}
\DeclarePairedDelimiter\ket{\lvert}{\rangle}
\DeclarePairedDelimiterX\braket[2]{\langle}{\rangle}{#1\,\delimsize\vert\,\mathopen{}#2}
\usepackage{pgfplots}
\pgfplotsset{compat=1.8}
\usetikzlibrary{arrows.meta}
\usepackage{tikz}
\usetikzlibrary{quantikz}
\usepackage{listings} 

\begin{document}

\title{Discrete Semantic States and Hamiltonian Dynamics in LLM Embedding Spaces}

\author{Timo\hspace*{1mm}Aukusti\hspace*{1mm}Laine \vspace*{0.5cm}}
\email{timo@financialphysicslab.com}

\begin{abstract}
We investigate the structure of Large Language Model (LLM) embedding spaces using mathematical concepts, particularly linear algebra and the Hamiltonian formalism, drawing inspiration from analogies with quantum mechanical systems. Motivated by the observation that LLM embeddings exhibit distinct states, suggesting discrete semantic representations, we explore the application of these mathematical tools to analyze semantic relationships. We demonstrate that the L2 normalization constraint, a characteristic of many LLM architectures, results in a structured embedding space suitable for analysis using a Hamiltonian formalism. We derive relationships between cosine similarity and perturbations of embedding vectors, and explore direct and indirect semantic transitions. Furthermore, we explore a quantum-inspired perspective, deriving an analogue of zero-point energy and discussing potential connections to Koopman-von Neumann mechanics.  While the interpretation warrants careful consideration, our results suggest that this approach offers a promising avenue for gaining deeper insights into LLMs and potentially informing new methods for mitigating hallucinations.
\end{abstract}

\maketitle

\section{Introduction}

Large Language Models (LLMs) have achieved significant success in natural language processing, becoming increasingly prevalent tools across various domains. However, the computational demands of these models present a substantial challenge, limiting their accessibility and widespread deployment. This article explores the structure of LLM embedding spaces, seeking to identify underlying principles that could lead to more efficient and reliable models.

We hypothesize that LLM embedding spaces exhibit a discrete structure, analogous to distinct states in physical systems, potentially mirroring aspects of quantum mechanical systems. This structure, characterized by specific relationships between embedding vectors, motivates the application of mathematical tools and methodologies.

Several compelling reasons motivate our exploration of connections between LLM embedding spaces and quantum mechanics, analyzed through the lens of the Hamiltonian formalism. LLMs exhibit inherent indeterminacy, producing outputs with a degree of uncertainty that resonates with the probabilistic nature of quantum systems. Furthermore, the massive computational resources required for training and deploying LLMs necessitate exploring alternative computational paradigms, such as quantum algorithms and quantum computers, which offer the potential for exponential speedups. The prevalence of hallucinations in LLMs, which undermines their reliability, particularly in critical applications, underscores the need for a deeper understanding of the embedding space structure and the origins of these inaccuracies. The Hamiltonian formalism, traditionally used to describe the dynamics of physical systems, provides a powerful framework for analyzing the transformations and relationships within LLM embedding spaces, potentially revealing underlying principles that govern their behavior and contribute to these challenges. Addressing these issues of computational cost and hallucinations is crucial for realizing the full potential of LLMs.

This article investigates the potential for leveraging mathematical methods, including those inspired by quantum mechanics and embodied in the Hamiltonian formalism. If similarities between LLM embedding spaces and other systems are established, quantum-inspired algorithms may offer performance improvements in LLM training and inference. Furthermore, insights into the nature of uncertainty, drawing from concepts like zero-point energy, might provide new strategies for managing and mitigating hallucinations, thereby enhancing the reliability and trustworthiness of LLMs.

We begin by examining the discrete structure of LLM embedding spaces, demonstrating how L2 normalization and other architectural features impose constraints on the relationships between embedding vectors. We consider a system of two distinct embedding vectors in the semantic space, which we term the LLM Embedding System. We then explore the mathematical formalisms used to describe this system, identifying key properties and exploring its characteristics. Finally, we discuss the implications of these findings for addressing hallucinations, including potential connections to quantum-inspired approaches.

\section{Background and Related Work}

Large Language Models (LLMs), such as those based on the Transformer architecture \cite{ref_vaswani}, have revolutionized natural language processing by utilizing high-dimensional embedding spaces to represent words, phrases, and sentences as numerical vectors. The dimensionality of these spaces typically ranges from hundreds to thousands of dimensions. The training process, often involving massive datasets and self-supervised learning objectives \cite{ref_bengio, ref_radford_gpt3}, shapes the geometry of these embedding spaces, encoding complex semantic relationships between linguistic units. The nature of these relationships and the structure of the resulting embedding spaces are active areas of research, with investigations focusing on understanding how semantic information is organized and accessed within these models \cite{ref_geva}.  The capabilities of these models, such as few-shot learning, have also been extensively studied \cite{ref_brown}.

Numerous studies have investigated the properties of LLM embedding spaces. A fundamental observation is that semantically similar words tend to be located proximally within the embedding space, as quantified by cosine similarity and other distance metrics \cite{ref_mikolov, ref_pennington}. This proximity reflects the model's ability to learn semantic relationships directly from data. Furthermore, researchers have explored how specific directions within the embedding space can be associated with particular semantic features or relationships, such as gender or sentiment \cite{ref_bolukbasi, ref_zhao}. Techniques like recursive matrix-vector spaces have also been used to capture semantic compositionality, enabling models to understand the meaning of phrases and sentences based on the meanings of their constituent words \cite{ref_socher}.

L2 normalization of embedding vectors is a common practice in LLMs, significantly influencing the geometry of the embedding space. Normalization encourages embeddings to reside on a hypersphere, which can simplify computations and enhance training stability. This constraint also affects the relationships between embedding vectors, as explored in this article. The impact of normalization on downstream task performance and the optimization landscape has been studied, with findings suggesting that normalization can improve generalization and prevent overfitting \cite{ref_arora}.

Various techniques have been employed to analyze the structure of LLM embedding spaces. Dimensionality reduction techniques, such as Principal Component Analysis (PCA) and t-distributed Stochastic Neighbor Embedding (t-SNE), have been used to visualize high-dimensional embedding spaces in lower dimensions, revealing clusters and patterns of semantic relationships \cite{ref_maalen}. Probing tasks, which involve training classifiers to predict properties of input text based on its embedding vector, provide insights into the information encoded within the embedding space \cite{ref_tenney}. Geometric analysis, involving the analysis of geometric properties such as the distribution of distances between embedding vectors and the curvature of the space, is also utilized to understand the underlying structure of these spaces \cite{ref_arora}.

The application of mathematical concepts to natural language processing is an area of growing interest. Models have been proposed for various NLP tasks, including text classification and information retrieval. These models often leverage principles such as superposition and entanglement to represent and process linguistic information. Information geometry, which studies statistical manifolds, has also been applied to analyze the structure of language models \cite{ref_amari}. Dynamical systems theory provides another lens for understanding the evolution of language models during training and inference \cite{ref_rabinovich}.

In previous work \cite{ref_laine1, ref_laine2, ref_laine3}, we have explored the analogy between LLM embedding spaces and mathematical concepts, positing that LLMs operate within a structured semantic space. Specifically, we introduced the concept of semantic wave functions to capture nuanced semantic interference effects \cite{ref_laine1}, clarified the core assumptions of a model for LLMs \cite{ref_laine2}, and demonstrated the feasibility of estimating semantic similarity using hardware, including the experimental calculation of cosine similarity \cite{ref_laine3}. These papers establish a foundation for the present work by highlighting the potential of mathematical principles to offer new perspectives on semantic representation and processing.

The potential of quantum computing for NLP tasks is also being explored. While the practical realization of quantum algorithms on current quantum hardware remains a challenge, classical algorithms inspired by quantum mechanics are being developed and applied to NLP. These quantum-inspired approaches offer alternative ways to represent and process linguistic information \cite{ref_wittek}.

It's important to acknowledge research focusing on the limitations of embedding spaces. While powerful, they can exhibit biases and fail to capture nuanced semantic relationships \cite{ref_hovy}. For example, Caliskan et al. (2017) demonstrated that semantics derived automatically from language corpora can contain human-like biases, reflecting societal stereotypes and prejudices \cite{ref_caliskan}. Therefore, any mathematical framework applied to these spaces must be interpreted with caution, considering the inherent limitations of the underlying representations.

\section{LLM Embedding Spaces Structure}
\label{sec_llm_embedding_spaces_structure}

In this section, we introduce an LLM Embedding System and demonstrate its discrete structure. We consider two distinct embedding vectors within a semantic space, representing a system where the vectors exhibit semantic dissimilarity.

\subsection{LLM Embedding System}

We model an LLM Embedding System, focusing on the relationship between dissimilar embedding vectors. Let $\mathbf{a}$ represent an arbitrary embedding vector

\begin{equation}
\mathbf{a} = \begin{bmatrix}
    a_1 \\
    a_2 \\
    \vdots \\
    a_N
\end{bmatrix}
\end{equation}

\noindent
For brevity, we represent this column vector as $\mathbf{a} = [a_{1}, a_{2},...,a_{N}]$, where the coefficients $a_i$ are real numbers. The vector $\mathbf{a}$ serves as a semantic anchor in the embedding space, allowing us to analyze the relationships with other vectors.

L2 normalization is a standard practice in LLM embedding spaces. We assume that $\mathbf{a}$ and all other vectors discussed in this section are L2-normalized, meaning they have a magnitude of one

\begin{equation}
    ||\mathbf{a}||^2 = \sum_{i=1}^Na_i^2  = 1 \label{eq_all_norm1}
\end{equation}

\begin{figure}[htpb]
    \centering
    \includegraphics[width=0.8\textwidth]{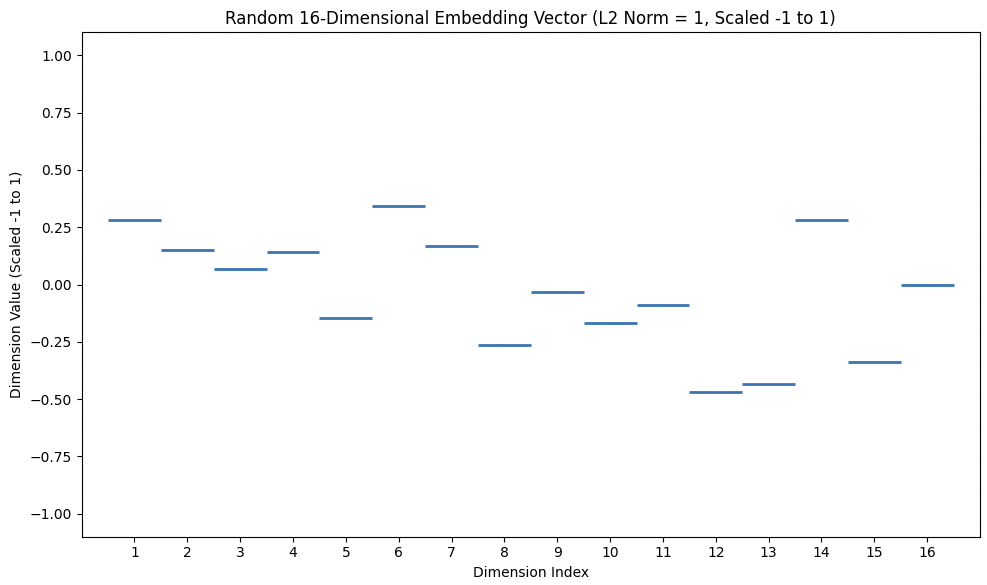}
\caption{Visualization of a 16-dimensional embedding vector, representing a semantic concept within an LLM's embedding space. The x-axis indicates the dimension index, while the y-axis represents the value of each dimension, scaled between -1 and 1. Although the specific values are arbitrary for illustrative purposes, the vector is L2-normalized, meaning its magnitude is 1. The horizontal lines depict the magnitude of each dimension, illustrating the discrete nature of the embedding vector components. This provides a visual representation of the structured nature of the embedding space.}
    \label{fig_states}
\end{figure}

\noindent
L2 normalization is not merely a mathematical convenience; it reflects the underlying structure of the embedding space induced by the LLM's training and architecture. Many LLMs utilize normalization layers or regularization techniques that encourage embeddings to lie on a hypersphere. This constraint helps to ensure that similarity is primarily determined by the angle between vectors, rather than their magnitudes. An example of an arbitrary embedding vector $\mathbf{a}$ is shown in Figure~\ref{fig_states}.

LLM embedding spaces are high-dimensional and encompass a vast number of potential token combinations. However, only a subset of these combinations yields meaningful semantic representations. Vector embeddings generated by models like Sentence Transformers aim to capture this meaningfulness by mapping semantically valid token combinations to specific regions within the space. We refer to vectors representing meaningful combinations as physical and those representing meaningless combinations as non-physical.

Now, consider a vector $\mathbf{b}$ in the same semantic space, defined as

\begin{equation}
    \mathbf{b} = [-a_{1}+\Delta_1, -a_{2}+\Delta_2 ,...,-a_{N}+\Delta_N]
\end{equation}

\noindent
where $\Delta_i$ represents a change in each dimension relative to the negated components of $\mathbf{a}$. If all $\Delta_i=0$, $\mathbf{b}$ becomes the maximally dissimilar vector to $\mathbf{a}$

\begin{equation}
    \mathbf{b}_{dis} = [-a_{1},- a_{2},...,-a_{N}]
\end{equation}

\noindent
The cosine similarity between $\mathbf{a}$ and $\mathbf{b}$ is defined as

\begin{equation}
    S_C(\mathbf{a},\mathbf{b})= \cos\theta = \frac{{\mathbf{a}\cdot \mathbf{b}}}{||\mathbf{a}|| \ ||\mathbf{b}||} \label{eq_cos_sim_real}
\end{equation}

\noindent
and for $\mathbf{b}_{dis}$, this simplifies to

\begin{equation}
    S_C(\mathbf{a},\mathbf{b}_{dis}) = \frac{\sum_{i=1}^N a_i(-a_i)}{||\mathbf{a}|| \ ||\mathbf{b}_{dis}||} =  -\sum_{i=1}^N a_i^2 = -1 \label{eq_cosine_sim}
\end{equation}

\noindent
Continuing the definition of our LLM Embedding System, we assume that the vectors $\mathbf{a}$ and $\mathbf{b}$ are neither perfectly similar ($S_C = 1$) nor perfectly dissimilar ($S_C = -1$). Perfect similarity would imply that we are dealing with the same vector, while perfect dissimilarity represents the opposite extreme. Instead, we focus on the more realistic scenario where $\mathbf{a}$ and $\mathbf{b}$ exhibit some degree of dissimilarity.
When only one $\Delta_i \ne 0$, we get degeneracies in other dimensions. To avoid this, an alternative way to define the LLM Embedding System is to consider embedding vectors whose dimensions are $k$ where $k$ is the amount of $\Delta_i$.

We posit the existence of a small positive value $\Delta > 0$ such that

\begin{equation}
    |\Delta_i| \ge |v_i|\Delta \label{eq_delta}
\end{equation}

\noindent
where $|v_i| \ge 1$ or $v_i = 0$. 
If $\Delta = 0$, then $S_C = -1$, which is an undesirable system state.
This means that the change in each dimension is bounded by a multiple of $\Delta$. Then we may write the cosine similarity as

\begin{equation}
    S_C(\mathbf{a}, \mathbf{b}) = {\mathbf{a}\cdot \mathbf{b}} = \sum_{i=1}^N a_i (-a_i+\Delta_i) = \sum_{i=1}^N a_i (-a_i+v_i\Delta) = -\sum_{i=1}^N a_i^2 + \Delta\sum_{i=1}^N v_i a_i = -1 + \Delta\sum_{i=1}^N v_i a_i > -1 \label{eq_cossim10}
\end{equation}

\noindent
This inequality holds if and only if the sum in Eq.~(\ref{eq_cossim10}) is greater than 0. And specifically,

\begin{equation}
    \sum_{i=1}^N v_ia_i  \ne 0 \label{eq_cossim11}
\end{equation}

\noindent
We will use this relation later. This condition implies that small changes in semantic features do not perfectly compensate for each other to maintain a constant overall similarity. While theoretically possible according to the cosine similarity definition, we assume this perfect compensation does not occur for physical embeddings. In other words, there isn't a multitude of subtly different vectors that all maintain the exact same similarity score with $\mathbf{a}$.

Furthermore, we assume that the embedding space is sufficiently dense, meaning that there are sufficiently many distinct states along each semantic dimension. This density is largely influenced by the amount of training data; more data generally leads to a denser space with more semantic states, while less data results in a sparser space. We also assume no degeneracy in the embedding space, meaning that $a_i \ne 0$ for all $i$. These are reasonable assumptions for physical embedding vectors, reflecting the richness and non-redundancy of semantic representations. Special cases will be handled separately.

We have now defined our LLM Embedding System and we conclude that it practically consists of two semantically different embedding vectors plus some assumptions which are reasonable for physical embedding vectors. In the next sections, we will examine the properties of this LLM Embedding System in more detail.

\subsection{Embedding Perturbations}

We now assume that $\mathbf{b}$ is a perturbation of the vector maximally dissimilar to $\mathbf{a}$

\begin{equation}
    \mathbf{b} = [-a_{1}+\Delta_1, -a_{2}+\Delta_2 ,...,-a_{N}+\Delta_N]
\end{equation}

\noindent
where the $\Delta_i$ represent changes in each dimension. These changes can be interpreted as semantic shifts or variations in meaning. Each dimension $i$ can be considered to represent a specific semantic feature or domain, with $a_i$ representing the strength of that feature in the embedding $\mathbf{a}$.

Given that both $\mathbf{a}$ and $\mathbf{b}$ are L2-normalized, the cosine similarity simplifies to the dot product

\begin{equation}
    S_C = \sum_{i=1}^N a_ib_i = \sum_{i=1}^N a_i(-a_i + \Delta_i) = \sum_{i=1}^N (-a_i^2 + a_i\Delta_i) = -\sum_{i=1}^N a_i^2 + \sum_{i=1}^N a_i\Delta_i
\end{equation}

\noindent
Since $\sum_{i=1}^N a_i^2 = 1$, we have

\begin{equation}
    S_C = -1 + \sum_{i=1}^N a_i\Delta_i \label{eq_equation5}
\end{equation}

\noindent
Because the cosine similarity ranges from -1 to 1, we have

\begin{equation}
    -1 \le -1 + \sum_{i=1}^N a_i\Delta_i \le 1
\end{equation}

\noindent
However, we must also enforce the constraint that $\mathbf{b}$ remains L2-normalized. This is essential because it reflects the inherent properties of LLM embedding spaces. This leads to the following condition:

\begin{equation}
    \sum_{i=1}^N b_i^2 = \sum_{i=1}^N (-a_i + \Delta_i)^2 = \sum_{i=1}^N (a_i^2 - 2a_i\Delta_i + \Delta_i^2) = \sum_{i=1}^N a_i^2 - 2\sum_{i=1}^N a_i\Delta_i + \sum_{i=1}^N \Delta_i^2 = 1
\end{equation}

\noindent
This implies that

\begin{equation}
	 -2\sum_{i=1}^N a_i\Delta_i + \sum_{i=1}^N \Delta_i^2 = 0 \label{eq_equation2}
\end{equation}

\noindent
This constraint can be interpreted as a form of "energy" conservation within the embedding space. Given that the L2 norm represents a measure of the embedding's magnitude, the constraint ensures that semantic shifts do not alter the overall magnitude of the representation. Geometrically, this constraint restricts the perturbation vector $\Delta$ to lie on a specific surface that is orthogonal to the vector $-2\mathbf{a} + \Delta$.

From equation (\ref{eq_equation2}) we obtain

\begin{equation}
    \sum_{i=1}^N a_i\Delta_i = \frac{1}{2}\sum_{i=1}^N \Delta_i^2
\end{equation}

\noindent
Substituting this back into the equation for the cosine similarity Eq.~(\ref{eq_equation5}), we obtain

\begin{equation}
    S_C = -1 + \frac{1}{2}\sum_{i=1}^N \Delta_i^2 \label{eq_equation6}
\end{equation}

\noindent
This result demonstrates that the cosine similarity between the original embedding vector $\mathbf{a}$ and its perturbed version $\mathbf{b}$ is equal to minus one plus half of the sum of the squared changes in each dimension. This relationship is a direct consequence of the L2 normalization of the embeddings and the constraint imposed on the perturbations.

\subsection{Smallest Semantic Perturbation}

Having established the relationship between cosine similarity and perturbations within the embedding space, we now explore the implications for the structured nature of this space. We propose that the discrete nature of language, as represented by tokens, induces a non-continuous structure within the embedding space.

Let us now investigate the scenario where vector $\mathbf{b}$ represents the smallest possible semantic perturbation of the vector maximally dissimilar to vector $\mathbf{a}$. This implies that their cosine similarity is slightly greater than -1

\begin{equation}
    S_{C1} = -1 + \epsilon_1
\end{equation}

\noindent
where $\epsilon_1$ is a small, non-zero positive constant. Using Eq.~(\ref{eq_equation6}), we obtain

\begin{equation}
    \epsilon_1 = \frac{1}{2}\sum_{i=1}^N \Delta_i^2
\end{equation}

\noindent
To minimize $\epsilon_1$ (i.e., to find the smallest semantic perturbation), we seek to minimize the sum of squared perturbations. The smallest possible change occurs when only one of the $\Delta_i$ is non-zero, with all others being zero. Let us assume that this non-zero perturbation occurs in the $i$-th dimension

\begin{equation}
    \epsilon_1 = \frac{1}{2}\Delta_i^2
\end{equation}

\noindent
This implies that the closest distinct vector $\mathbf{b}$ differs from $\mathbf{a}$ by a small amount $\Delta_i$ in a single dimension:

\begin{align}
    \mathbf{a} &= [a_{1}, a_{2},...,a_i,...,a_{N}] \\
     \mathbf{b} &= [-a_{1}, -a_{2},...,-a_{i}+\Delta_i,...,-a_{N}] \label{eq_vectorb}
\end{align}

\noindent
The value of $\Delta_i$, constrained by training and architecture, reflects the discrete nature of LLM tokenization and the limitations of floating-point representation. This is analogous to quantum mechanics, where boundary conditions lead to quantization. The smallest semantic change is therefore a discrete, non-infinitesimal difference between embeddings.

Furthermore, if there is only one $\Delta_i$ as a perturbation, then according to Eq.~(\ref{eq_equation2}) its value must be

\begin{equation}
\Delta_i = 2a_i \label{eq_delta_equation}
\end{equation}

\noindent
To demonstrate this, consider Equation (\ref{eq_equation2})

\begin{equation}
-2\sum_{i=1}^N a_i\Delta_i + \sum_{i=1}^N \Delta_i^2 = 0
\end{equation}

\noindent
If only one $\Delta_i$ is non-zero, then the equation simplifies to

\begin{equation}
-2 a_i \Delta_i + \Delta_i^2 = 0
\end{equation}

\noindent
Solving for $\Delta_i$, we get

\begin{equation}
\Delta_i (2 a_i -\Delta_i) = 0
\end{equation}

\noindent
The solutions are $\Delta_i = 0$ (which corresponds to no perturbation) and $\Delta_i = 2a_i$. We obtain another constraint: if $\Delta_i$ represents the smallest change in the embedding space, then, according to Eq.~(\ref{eq_delta_equation}), the change must occur in the dimension with the smallest absolute value $|a_i|$. If this were not the case, and we considered some $a_j$ where $|a_j| > |a_i|$, then the corresponding $\Delta_j$ would not represent the smallest possible perturbation, contradicting our initial requirement. However, this analysis does not restrict the sign of the smallest $a_i$.

We can now rewrite the embedding vector $\mathbf{b}$ from Eq.~(\ref{eq_vectorb}) as

\begin{equation}
     \mathbf{b} = [-a_{1}, -a_{2},...,a_{i},...,-a_{N}] \label{eq_vectorb2}
\end{equation}

\noindent
and the cosine similarity for the first perturbed state is

\begin{equation}
    S_{C1} = -1 + 2a_i^2 \label{eq_firststate}
\end{equation}

\begin{figure}[htpb]
    \centering
    \includegraphics[width=0.8\textwidth]{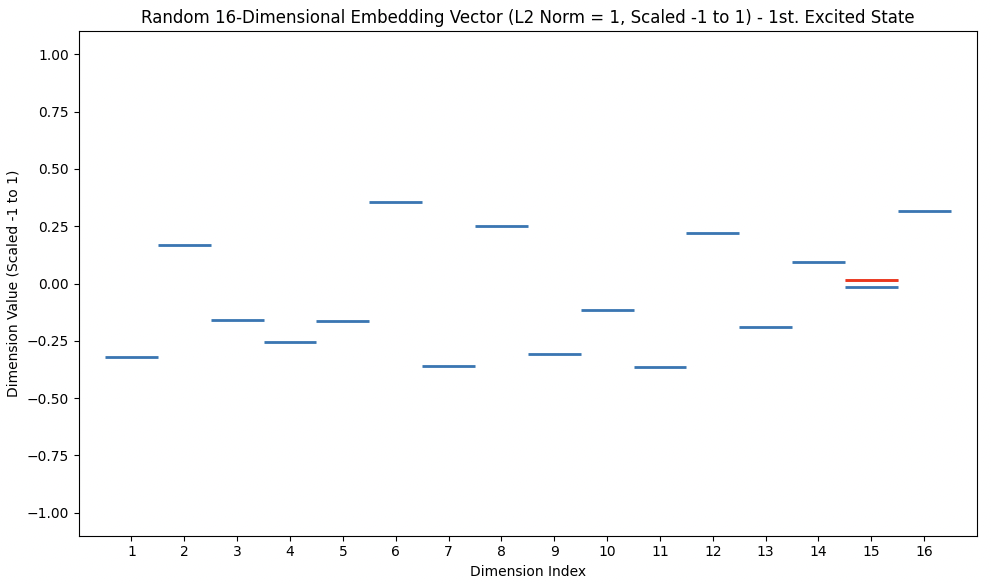}
\caption{Visualization of a 16-dimensional embedding vector representing a semantic concept. Horizontal lines show dimension magnitudes (scaled -1 to 1). A red line indicates the first perturbed state, where the value of one dimension is inverted compared to the fully dissimilar state, illustrating the smallest possible semantic change.}
    \label{fig_excited_state}
\end{figure}

\noindent
As an example of the first perturbed state is shown in Figure~\ref{fig_excited_state} with the red color.

This result implies that, within the constraints of L2 normalization and minimizing the L2 norm of the perturbation, a change in cosine similarity corresponds to inverting the value of one dimension. This further implies that if we consider an LLM embedding vector in semantic space and examine the smallest change in cosine similarity of its maximally dissimilar vector, we find that the value of one dimension changes sign. This means that if we have some semantic context or sentence, and a dimension represents a semantic feature or a domain, then the smallest change in one semantic feature corresponds to inverting that feature.

\subsection{Higher Perturbations}

To further explore the structure of the embedding space, we now consider higher-order perturbations. Specifically, we define the cosine similarity for a second perturbed state as

\begin{equation}
    S_{C2} = -1 + \epsilon_2
\end{equation}

\noindent
where $S_{C2}$ represents the cosine similarity between the original and dissimilar perturbed embedding vectors, and $\epsilon_2$ is a small positive constant representing the excitation of this state. We assume that $\epsilon_2 > \epsilon_1$.

Let us consider perturbing two dimensions of the original embedding vector $\mathbf{a}$

\begin{align}
    \mathbf{a} &= [a_{1}, a_{2},...,a_i,...,a_j,...,a_{N}] \\
     \mathbf{b} &= [-a_{1}, -a_{2},...,-a_{i}+\Delta_i,...,-a_{j}+\Delta_j,...,-a_{N}]
\end{align}

\noindent
Alternatively, we could consider changing the same dimension by a larger amount

\begin{align}
    \mathbf{a} &= [a_{1}, a_{2},...,a_i,...,a_{N}] \\
     \mathbf{b}' &= [-a_{1}, -a_{2},...,-a_{i}+\Delta_{i1}+\Delta_{i2},...,-a_{N}]
\end{align}

\noindent
Let us focus on vectors $\mathbf{a}$ and $\mathbf{b}$. From Eq.~(\ref{eq_equation2}), we obtain

\begin{equation}
    -2 (a_i\Delta_i + a_j\Delta_j) + (\Delta_i^2+\Delta_j^2) = 0 \label{eq_contraint1}
\end{equation}

\noindent
We now use the relation Eq.~(\ref{eq_delta}) and write Eq.~(\ref{eq_contraint1}) as

\begin{equation}
    -2 \Delta(v_ia_i + v_ja_j) + \Delta^2(v_i^2+v_j^2) = 0 \label{eq_contraint2}
\end{equation}

\noindent
We may solve

\begin{align}
    \Delta_i &=  \frac{2v_i}{v_i^2+v_j^2}(v_ia_i + v_ja_j) \\
    \Delta_j &=  \frac{2v_j}{v_i^2+v_j^2}(v_ia_i + v_ja_j)
\end{align}

\noindent
Now using Eq.~(\ref{eq_cossim11}) we can first conclude that both $\Delta_i \ne 0$ and $\Delta_j \ne 0$. And if setting $v_i = 1$ and $v_j = 0$, the equation reduces to Eq.~(\ref{eq_delta_equation}). The cosine similarity is

\begin{equation}
 S_{C2} = -1+\frac{2}{v_i^2+v_j^2}(v_ia_i + v_ja_j)^2
\end{equation}

\noindent
Now depending on the values of $a_i$ and $a_j$, $S_{C2}$ can be smaller or larger than $S_{C1}$, but at least it is greater than $-1$. A special case is obtained, for example, when $v_i = v_j =1$. Then the cosine similarity reduces to

\begin{equation}
 S_{C2} = -1+(a_i + a_j)^2
\end{equation}

\noindent
and the perturbed embedding vector $\mathbf{b}$ becomes

\begin{equation}
     \mathbf{b} = [-a_{1}, -a_{2},...,a_{j},...,a_{i},...,-a_{N}] \label{eq_exct21}
\end{equation}

\begin{figure}[htpb]
    \centering
    \includegraphics[width=0.8\textwidth]{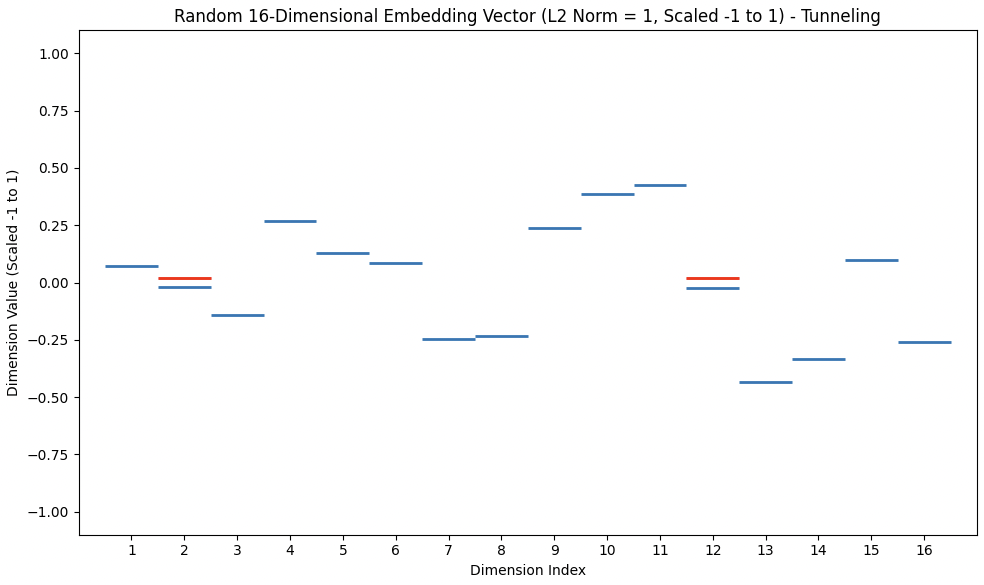}
\caption{IIlustration of the embedding vector $\mathbf{b}$ after a specific perturbation. Horizontal lines represent the values of each dimension in the vector, scaled from -1 to 1. Red lines highlight dimensions $j$ and $i$, which have undergone similar perturbations, effectively swapping their original values. This type of perturbation corresponds to a transition to a higher excited state in the LLM embedding space.}
    \label{fig_nextstate}
\end{figure}

\noindent
A visualization of this case is shown in Figure~\ref{fig_nextstate}.

Interestingly, in this specific scenario, components $a_i$ and $a_j$ effectively exchange positions. This suggests that, under these specific constraints, the next excited state can be interpreted as a swapping of the semantic roles or importances of the $i$-th and $j$-th features within the embedding. This could correspond to a semantic transformation where the emphasis shifts from one concept to another related, but distinct, concept. For example, if $a_i$ represents the feature "color" and $a_j$ represents the feature "size", this transformation could represent a shift from emphasizing the color of an object to emphasizing its size.

One can continue to analyze higher-order perturbations, both analytically and numerically. For example, when considering the general case

\begin{equation}
    \mathbf{b} = [-a_{1}+\Delta_1, -a_{2}+\Delta_2 ,...,-a_{N}+\Delta_N] \label{eq_alldelta}
\end{equation}

\noindent
we get a cosine similarity

\begin{equation}
 S_{CN} = -1+\frac{2}{\sum_{i=1}^N v_i^2}\left (\sum_{i=1}^N v_ia_i\right )^2 \label{eq_cossimall3}
\end{equation}

\noindent
One can numerically verify that the Eq.~(\ref{eq_cossimall3}) holds.
When generating embedding vectors using tools like Sentence Transformers, it's very common for the two vectors to have different values across all dimensions.
So Eqs.~(\ref{eq_alldelta}) and (\ref{eq_cossimall3}) are common forms for a physical embedding vector.

If we further consider a third embedding vector like

\begin{equation}
    \mathbf{c} = [ -a_{1}+\Delta_1', -a_{2}+\Delta_2' ,...,-a_{N}+\Delta_N'] \label{eq_alldelta2}
\end{equation}

\noindent
we may use the same results derived earlier and redefine $\Delta_i$s, like

\begin{equation}
  \tilde{\Delta}_i = \Delta_i + \Delta_i'
\end{equation}

\section{Hamiltonian Representation of Semantic Transformations}
\label{hamiltonian_representation}

We further investigate the cosine similarity by introducing a Hamiltonian matrix to represent the transformation between embedding vectors. This approach enables us to leverage tools from linear algebra to analyze the structure of the embedding space. Vector $\mathbf{b}$ can be expressed as

\begin{equation}
    \mathbf{b} = \mathbf{H} \mathbf{a}
\end{equation}

\noindent
where

\begin{equation}
   \begin{bmatrix}
        b_1 \\
        b_2 \\
        \vdots \\
        b_N
    \end{bmatrix} =
    \begin{bmatrix}
        a_1 +\Delta_1 \\
        a_2 +\Delta_2\\
        \vdots \\
        a_N +\Delta_N
    \end{bmatrix}  =
   \begin{bmatrix}
        h_{11} & h_{12} & \cdots & h_{1N} \\
        h_{21} & h_{22} & \cdots & h_{2N} \\
        \vdots & \vdots & \ddots & \vdots \\
        h_{N1} & h_{N2} & \cdots & h_{NN}
    \end{bmatrix}
  \begin{bmatrix}
        a_1 \\
        a_2 \\
        \vdots \\
        a_N
    \end{bmatrix},
\end{equation}

\noindent
Given that both $\mathbf{a}$ and $\mathbf{b}$ are L2 normalized, the cosine similarity can be expressed as

\begin{equation}
    S_C = \mathbf{a}^T \mathbf{H} \mathbf{a} \label{eq_equation9}
\end{equation}

\noindent
This provides a compact and insightful representation of the relationship between embedding vectors. It allows us to view the transformation from $\mathbf{a}$ to $\mathbf{b}$ as a linear operation, which can be analyzed in terms of the eigenvalues and eigenvectors of the Hamiltonian matrix.

\subsection{Types of Semantic Transitions}

We can identify three distinct cases for the transformation between embedding vectors.

\subsubsection{Same or Maximally Dissimilar Embedding Vectors}

The simplest case occurs when $\mathbf{b} = \mathbf{a}$. In this scenario, $\mathbf{H}$ is the identity matrix, denoted by $\mathbf{I}$, where $h_{ii} = 1$ for all $i$ and $h_{ij} = 0$ for $i \neq j$. This corresponds to no transformation, and the cosine similarity is 1. Similarly, if $\mathbf{b} = -\mathbf{a}$, then $\mathbf{H} = -\mathbf{I}$.

\subsubsection{Direct Transitions}

Let us now consider the case where $\mathbf{a}$ and $\mathbf{b}$ are distinct embedding vectors. A direct transition implies that each component $b_i$ of vector $\mathbf{b}$ depends only on the corresponding component $a_i$ of vector $\mathbf{a}$. Mathematically, this can be expressed as

\begin{equation}
    b_i = h_{ii} a_i
\end{equation}

\noindent
with $h_{ij} = 0$ when $i \ne j$. This represents a scenario where the transformation from $\mathbf{a}$ to $\mathbf{b}$ involves only scaling the individual dimensions of $\mathbf{a}$. This can be interpreted as a transformation that changes the strength of individual semantic features without introducing any dependencies between them. The Hamiltonian $\mathbf{H}$ is a diagonal matrix in this case, and the diagonal elements, $h_{ii}$, represent the scaling factors for each semantic feature.

\subsubsection{Indirect Transitions}

Let us now consider three distinct embedding vectors, representing meaningful semantic states, which we denote as $\ket{1}$, $\ket{2}$, and $\ket{3}$. These vectors can be interpreted as embedding vectors corresponding to specific token combinations that possess semantic coherence. The use of Dirac notation (ket notation) is intended to draw an analogy to quantum mechanics and to facilitate the application of quantum mechanical tools for analyzing LLM embedding spaces. Our objective is to understand how transitions between these states can be achieved and how indirect transitions relate to direct transitions.

A direct transformation from state $\ket{1}$ to state $\ket{2}$ can be achieved by applying a linear operator, $\mathbf{H}_{1\rightarrow2}$. The same principle applies to direct transformations between any two of these states (e.g., $\ket{1}$ to $\ket{3}$ or $\ket{2}$ to $\ket{3}$). Mathematically, this can be expressed as

\begin{equation}
\ket{3} = \mathbf{H}_{1\rightarrow3} \ket{1}
\end{equation}

\noindent
where $\mathbf{H}_{1\rightarrow3}$ is a Hamiltonian representing the direct transition from state $\ket{1}$ to state $\ket{3}$. This direct transition represents a relationship between semantic states. For example, $\ket{1}$ might represent the concept "quick brown fox," and $\ket{3}$ might represent "a fast brown fox." The Hamiltonian $\mathbf{H}_{1\rightarrow3}$ transforms the semantic features in $\ket{1}$ to align with those in $\ket{3}$.

\begin{figure}[htpb]
    \centering
    \includegraphics[width=0.7\textwidth]{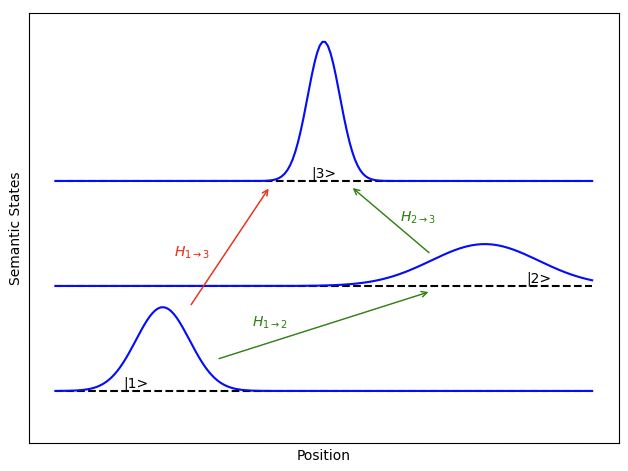}
\caption{Quantum-inspired representation of semantic relationships in LLM embedding space. Three semantic states, $\ket{1}$, $\ket{2}$, and $\ket{3}$, are visualized as energy levels with associated population distributions, drawing an analogy to quantum mechanical systems. The arrows represent effective Hamiltonians governing transitions between these states, with $H_{1 \rightarrow 3}$ indicating a direct transformation from $\ket{1}$ to $\ket{3}$, and $H_{1 \rightarrow 2}$ and $H_{2 \rightarrow 3}$ representing sequential transformations through the intermediate state $\ket{2}$, suggesting a more complex semantic relationship.}
    \label{fig_3levels}
\end{figure}

However, a key insight arises when we consider indirect transitions. Suppose we want to transform state $\ket{1}$ to state $\ket{3}$, but we do so through the intermediate state $\ket{2}$. This involves two sequential transformations: first, from $\ket{1}$ to $\ket{2}$, and then from $\ket{2}$ to $\ket{3}$. This indirect transition may suggest a more complex semantic relationship between $\ket{1}$ and $\ket{3}$. For example, $\ket{1}$ might represent the concept "quick brown fox," $\ket{2}$ might represent "a fast brown animal," and $\ket{3}$ might represent "a fast brown fox." The intermediate state $\ket{2}$ introduces a level of abstraction that is not present in the direct transition from $\ket{1}$ to $\ket{3}$,  see the Figure~\ref{fig_3levels}.

The first transformation, from $\ket{1}$ to $\ket{2}$, is represented by a Hamiltonian $\mathbf{H}_{1\rightarrow2}$. We can then determine the vector $\ket{2}$, which represents the intermediate state, by applying the Hamiltonian $\mathbf{H}_{1\rightarrow2}$ to the vector $\ket{1}$

\begin{equation}
\ket{2} = \mathbf{H}_{1\rightarrow2} \ket{1}
\end{equation}

\noindent
This results in a new vector, which we denote as $\ket{2}$.
The second transformation, from $\ket{2}$ to $\ket{3}$, can then be represented relative to the first transformation.
To express this mathematically, let us represent the transformation from $\ket{2}$ to $\ket{3}$ as

\begin{equation}
\ket{3} = \mathbf{H}_{2\rightarrow3} \ket{2}
\end{equation}

\noindent
Applying this Hamiltonian to the intermediate state $\ket{2}$ yields the final state $\ket{3}$

\begin{equation}
\ket{3} = \mathbf{H}_{2\rightarrow3} \ket{2}
\end{equation}

\noindent
To analyze the transformations in a common basis, we can perform a change of basis. Let $\mathbf{U}$ be a unitary transformation that changes the basis from $\ket{2}$ to $\ket{1}$. We can then express $\ket{3}$ in terms of the basis of $\ket{1}$

\begin{equation}
\ket{3} = \mathbf{H}_{2\rightarrow3} \mathbf{H}_{1\rightarrow2} \ket{1}
\end{equation}

\noindent
It is important to note that $\mathbf{H}_{2\rightarrow3}$ is expressed in the basis of $\ket{2}$. To express this in the basis of $\ket{1}$, we can use the unitary transformation to obtain

\begin{equation}
\mathbf{H'}_{2\rightarrow3} = \mathbf{U} \mathbf{H}_{2\rightarrow3} \mathbf{U}^{\dagger}
\end{equation}

\noindent
where the new Hamiltonian $\mathbf{H'}_{2\rightarrow3}$ represents the transformation from $\ket{2}$ to $\ket{3}$ in the basis of $\ket{1}$.
Applying this transformed Hamiltonian to $\mathbf{H}_{1\rightarrow2}\ket{1}$ yields the final state $\ket{3}$

\begin{equation}
\ket{3} = \mathbf{H'}_{2\rightarrow3} \mathbf{H}_{1\rightarrow2}\ket{1}
\end{equation}

\noindent
The overall transformation from $\ket{1}$ to $\ket{3}$ through $\ket{2}$ is then given by

\begin{equation}
\ket{3} = \mathbf{H}_{\text{indirect}} \ket{1}
\end{equation}

\noindent
where $\mathbf{H}_{\text{indirect}} = \mathbf{H'}_{2\rightarrow3} \mathbf{H}_{1\rightarrow2}$ is the effective Hamiltonian for the indirect transition. Generally, the indirect Hamiltonian has off-diagonal elements.

In a realistic LLM embedding space, there are likely multiple pathways to transition from state $\ket{1}$ to state $\ket{3}$. Each pathway corresponds to a different intermediate state (or sequence of intermediate states) and, therefore, to a different effective Hamiltonian. This suggests a superposition of possible transition pathways. The structure of the effective Hamiltonian, specifically the presence and magnitude of off-diagonal elements, reflects the complexity of the semantic relationship between the initial and final states. A diagonal Hamiltonian indicates a direct, straightforward relationship, while a Hamiltonian with significant off-diagonal elements suggests a more indirect and nuanced relationship mediated by the intermediate states. The analysis of these effective Hamiltonians could provide insights into how LLMs process and represent complex semantic relationships. The relative contributions of different pathways could be related to the attention mechanisms within the Transformer architecture, with attention weights potentially reflecting the probability amplitudes associated with each pathway. This framework provides a bridge between the abstract mathematical representation of semantic transitions and the concrete mechanisms of LLM operation.

We note that because Eq.~(\ref{eq_equation9}) expresses the cosine similarity as a quadratic form, even if the Hamiltonian $\mathbf{H}$ is not symmetric, we can always replace it with a symmetric Hamiltonian that yields the same cosine similarity value. This means that, for the purpose of calculating cosine similarity, the direction of the transition between embedding vectors does not fundamentally alter the result. A symmetric Hamiltonian implies that the relationship between semantic states is reciprocal, while a non-symmetric Hamiltonian could capture directional dependencies. The fact that the cosine similarity remains the same does not imply that the Hamiltonians are identical. The choice of using a symmetric or non-symmetric Hamiltonian should be guided by the specific research question and the desire to capture either reciprocal or directional relationships between semantic states.

\subsection{Constraint on Hamiltonian Coefficients}

The L2 normalization of the embedding vectors imposes a constraint on the elements of the Hamiltonian matrix $\mathbf{H}$. This constraint arises from the requirement that both $\mathbf{a}$ and $\mathbf{b}$ reside on the unit hypersphere. We can derive this constraint by relating the components of $\mathbf{b}$ to the components of $\mathbf{a}$ and the elements of $\mathbf{H}$

\begin{equation}
    b_i = a_i+\Delta_i = \sum_{j=1}^N h_{ij}a_j
\end{equation}

\noindent
From this relationship, we can express the perturbation $\Delta_i$ in terms of the elements of $\mathbf{H}$ and the components of $\mathbf{a}$

\begin{equation}
   \Delta_i = \sum_{j=1}^N h_{ij}a_j - a_i \label{eq_deltah1}
\end{equation}

\noindent
Squaring both sides of the above equation, we obtain

\begin{equation}
   \Delta_i^2 = \left( \sum_{j=1}^N h_{ij}a_j - a_i \right)^2
    = \left( \sum_{j=1}^N h_{ij}a_j \right)^2 - 2a_i\sum_{j=1}^N h_{ij}a_j + a_i^2 \label{eq_deltah2}
\end{equation}

\noindent
Given that both $\mathbf{a}$ and $\mathbf{b}$ are L2 normalized, we have

\begin{equation}
    \sum_{i=1}^N b_i^2 = \sum_{i=1}^N (a_i+\Delta_i)^2 = \sum_{i=1}^N (a_i^2+2a_i\Delta_i+\Delta_i^2) = 1 \label{eq_sumb}
\end{equation}

\noindent
Substituting Eq.~(\ref{eq_deltah1}) into Eq.~(\ref{eq_sumb}) to eliminate $\Delta_i$, we get

\begin{equation}
    \sum_{i=1}^N b_i^2 = \sum_{i=1}^N \left[ a_i^2+2a_i\left( \sum_{j=1}^N h_{ij}a_j-a_i \right)+ \left( \sum_{j=1}^N h_{ij}a_j\right)^2 - 2a_i\sum_{j=1}^N h_{ij}a_j + a_i^2 \right] \label{eq_sumb1}
\end{equation}

\noindent
Simplifying the above equation and using the fact that $\sum_{i=1}^N a_i^2 = 1$, we arrive at the following constraint on the Hamiltonian coefficients

\begin{equation}
    C = \sum_{i=1}^N  \left( \sum_{j=1}^N h_{ij}a_j \right)^2 = 1 \label{eq_sumb2}
\end{equation}

\noindent
This constraint ensures that the transformed vector $\mathbf{b} = \mathbf{H}\mathbf{a}$ remains L2 normalized. It implies that the Hamiltonian $\mathbf{H}$ cannot be an arbitrary matrix; its elements must satisfy this condition to preserve the geometry of the embedding space. This constraint could potentially be used to regularize the training of LLMs, encouraging them to learn embedding spaces that respect the L2 normalization property and discouraging transformations that would distort the hyperspherical geometry.

\subsection{Experimental Validation of the Hamiltonian Approach}

To illustrate the concepts discussed above, we present a numerical example using the Sentence Transformers embedding model "gemma-300m" with 768 dimensions. This experiment aims to demonstrate the validity of the Hamiltonian approach and to verify the constraint imposed by L2 normalization.

The procedure begins by defining three text prompts: "quick brown fox" ($\ket{1}$), "a fast brown animal" ($\ket{2}$), and "a fast brown fox" ($\ket{3}$). These prompts are encoded into embedding vectors using the Sentence Transformers model, and the resulting vectors (vec1, vec2, vec3) are automatically L2 normalized by the model.
A direct cosine similarity between vec1 and vec3 is then calculated as a baseline

\begin{equation}
S_{C}^{\text{direct}} (\ket{1}, \ket{3}) = 0.87610495
\end{equation}

\noindent
To model the indirect transition from $\ket{1}$ to $\ket{3}$ via $\ket{2}$, we first determine a unitary transformation $\mathbf{H}_{1\rightarrow2}$ that transforms $\ket{1}$ to $\ket{2}$. This transformation is calculated using the Householder transformation, which guarantees that $\mathbf{H}_{1\rightarrow2}$ is a unitary matrix, ensuring that the L2 norm is preserved. The resulting vector representing $\ket{2}$ is then obtained by applying the Hamiltonian to $\ket{1}$

\begin{equation}
\ket{2} = \mathbf{H}_{1\rightarrow2} \ket{1}
\end{equation}

\noindent
Next, we determine another unitary transformation $\mathbf{H}_{2\rightarrow3}$ that transforms $\ket{2}$ to $\ket{3}$. Again, this transformation is calculated using the Householder transformation. The resulting vector representing $\ket{3}$ via the indirect path is obtained by applying the Hamiltonian to $\ket{2}$

\begin{equation}
\ket{3}_{\text{indirect}} = \mathbf{H}_{2\rightarrow3} \ket{2}
\end{equation}

\noindent
The indirect cosine similarity is then calculated as

\begin{equation}
S_{C}^{\text{indirect}} (\ket{1}, \ket{3}_{\text{indirect}}) = 0.87610488
\end{equation}

\noindent
We also checked the constraint value for both $\mathbf{H}_{1\rightarrow2}$ and $\mathbf{H}_{2\rightarrow3}$, resulting in

\begin{align}
 C_{12} &= 1.00000000 \\
 C_{23} &= 1.00000003
\end{align}

\noindent
This confirms that the transformations remain L2 normalized. Within the limits of numerical accuracy, all equations are satisfied.

The results of this numerical experiment demonstrate that, through the use of unitary transformations, the direct and indirect cosine similarities are equal, and the constraint imposed by L2 normalization is satisfied. This provides numerical validation of the theoretical framework presented in this section, supporting the application of the Hamiltonian approach for analyzing semantic transitions in LLM embedding spaces.

\section{Classical Hamiltonian System}

We are now ready to combine results from sections \ref{sec_llm_embedding_spaces_structure} and \ref{hamiltonian_representation}.
We begin with the cosine similarity expressed in terms of a Hamiltonian matrix

\begin{equation}
S_C = \mathbf{a}^T \mathbf{H} \mathbf{a}
\end{equation}

\noindent
To explore potential relationships with bounded quantities, we transform the cosine similarity to a range between 0 and 1

\begin{equation}
S_C' = \frac{1}{2}(S_C + 1)
\end{equation}

\noindent
If $S_C$ ranges from -1 to 1, then $S_C'$ ranges from 0 to 1. This transformation allows us to express the similarity in a form that resembles probabilities or normalized measures.

We can express this transformed similarity using a modified Hamiltonian

\begin{equation}
S_C' = \mathbf{a}^T \mathbf{H'} \mathbf{a}
\end{equation}

\noindent
where

\begin{equation}
\mathbf{H'} = \frac{1}{2}(\mathbf{H} + \mathbf{I})
\end{equation}

\noindent
Here, $\mathbf{I}$ denotes the identity matrix. This transformation shifts the eigenvalues of the Hamiltonian but preserves its eigenvectors. Also, the value of $S_C'$ if from Eq.~(\ref{eq_cossimall3}) is

\begin{equation}
 S_{CN}' = \frac{1}{\sum_{i=1}^N v_i^2}\left (\sum_{i=1}^N v_ia_i\right )^2 \label{eq_cossimall4}
\end{equation}

\noindent
Note that $S_{CN}'$ is always positive and from Eq.~(\ref{eq_cossim11}) it follows that $S_{CN}' \ne 0$.

\subsection{First Perturbation and Parity Symmetry}

Using the notations defined earlier, the transformed cosine similarity for the first perturbed state Eq.~(\ref{eq_firststate}) is

\begin{equation}
  S_{C1}' = a_i^2
\end{equation}

\noindent
The corresponding Hamiltonian is

\begin{equation}
\mathbf{H'_1} =
\begin{bmatrix}
    0 & 0 & \cdots & 0 & \cdots & 0 \\
    0 & 0 & \cdots & 0 & \cdots & 0 \\
    \vdots & \vdots & \ddots & \vdots & \ddots & \vdots \\
    0 & 0 & \cdots & 1 & \cdots & 0 \\
    \vdots & \vdots & \ddots & \vdots & \ddots & \vdots \\
    0 & 0 & \cdots & 0 & \cdots & 0
\end{bmatrix}
\end{equation}

\noindent
where $H'_{ll} = 1$, if element is $ii$, and $0$ otherwise.
As one can see, $Tr(\mathbf{H'_1} )=1$. The trace of a square matrix is the sum of its diagonal elements. It is invariant under a change of basis and equals the sum of the matrix's eigenvalues, providing valuable information about the matrix's properties. In this case, the trace indicates that $\mathbf{H'_1}$ has one eigenvalue equal to 1 and all other eigenvalues equal to 0.

The effect of this first-order perturbation can be related to the concept of parity symmetry. We can define a parity operator $P$ that inverts the sign of one or more dimensions of the embedding vector $\mathbf{a}$. For example, inverting the $i$-th dimension yields

\begin{equation}
P =
\begin{bmatrix}
1 & 0 & 0 & \cdots & 0 \\
0 & 1 & 0 & \cdots & 0 \\
0 & 0 & -1 & \cdots & 0 \\
\vdots & \vdots & \vdots & \ddots & \vdots \\
0 & 0 & 0 & \cdots & 1
\end{bmatrix}
\end{equation}

\noindent
Applying $P$ to $\mathbf{a}$ yields a parity-transformed vector $\mathbf{a'}$

\begin{equation}
\mathbf{a'} = P \mathbf{a}
\end{equation}

\noindent
If a minimal perturbation involves inverting the sign of one dimension, we can relate this to a parity transformation.
If the LLM Embedding System exhibits parity symmetry, the semantic coherence of $\mathbf{a'}$ should be similar to that of $\mathbf{a}$, implying a form of invariance under the parity operation. This would suggest that the expectation value of the original Hamiltonian $H$ should be the same for both $\mathbf{a}$ and $\mathbf{a'}$

\begin{equation}
\langle \mathbf{a'}|H|\mathbf{a'} \rangle = \langle \mathbf{a}|H|\mathbf{a} \rangle
\end{equation}

\noindent
This suggests that inverting a dimension corresponds to negating a semantic feature. If a concept and its negation are equally coherent (e.g., positive versus negative sentiment), the system exhibits parity symmetry. The sign inversion is related to the eigenvalues of $P$ (+1 and -1). The parity-transformed state is associated with the eigenvalue -1, leading to the sign inversion. However, it is important to note that $\mathbf{H'_1}$ itself does not perform this parity transformation; it represents the effect of a minimal perturbation, which can be related to the concept of parity symmetry.

\subsection{Higher Perturbations and Rotational Symmetry}

As an example for higher perturbations, we consider the specific case where $v_i = v_j = 1$. This leads to

\begin{equation}
  S_{C2}' = \frac{1}{2}(a_i+a_j)^2
\end{equation}

\noindent
The corresponding Hamiltonian is

\begin{equation}
\mathbf{H'_2} = \frac{1}{2}
\begin{bmatrix}
    0 & 0 & \cdots & 0 & 0 & \cdots & 0 \\
    0 & 0 & \cdots & 0 & 0 & \cdots & 0 \\
    \vdots & \vdots & \ddots & \vdots  & \vdots & \ddots & \vdots \\
    0 & 0 & \cdots & 1 & 1 & \cdots & 0 \\
     0 & 0 & \cdots & 1 & 1  & \cdots & 0 \\
    \vdots & \vdots & \ddots & \vdots  & \vdots & \ddots & \vdots \\
    0 & 0 & \cdots & 0 & 0 & \cdots & 0
\end{bmatrix}
\end{equation}

\noindent
where $H'_{ll} = 1/2$, if element is $ii, jj, ij$ or $ji$, and $0$ otherwise.
As before, $Tr(\mathbf{H'_2} )=1$. This signifies that the total probability is conserved, implying that the system remains within the defined phase space and no particles are being created or destroyed, a consequence of Liouville's theorem.

The form of this Hamiltonian suggests a possible link to rotational symmetry in the $i$-$j$ plane. To analyze rotational symmetry, we define a rotation operator $R(\theta)$ that transforms the embedding vector $\mathbf{a}$ into $\mathbf{a'}$

\begin{align}
a'_i &= a_i \cos(\theta) - a_j \sin(\theta) \\
a'_j &= a_i \sin(\theta) + a_j \cos(\theta)\\
a'_k &= a_k \quad \text{for } k \neq i, j
\end{align}

\noindent
where $a_i$ and $a_j$ are components of $\mathbf{a}$, and $a'_i$ and $a'_j$ are components of $\mathbf{a'}$.

If the LLM Embedding System exhibits rotational symmetry in the $i$-$j$ plane, the semantic coherence of $\mathbf{a'}$ should be similar to that of $\mathbf{a}$, implying a form of invariance under the rotation operation. This would suggest that the expectation value of the original Hamiltonian $H$ should be the same for both $\mathbf{a}$ and $\mathbf{a'}$

\begin{equation}
\langle \mathbf{a'}|H|\mathbf{a'} \rangle = \langle \mathbf{a}|H|\mathbf{a} \rangle
\end{equation}

\noindent
Interestingly, the form of the perturbed vector $\mathbf{b}$ derived earlier

\begin{equation}
     \mathbf{b} = [-a_{1}, -a_{2},...,a_{j},...,a_{i},...,-a_{N}]
\end{equation}

\noindent
exhibits characteristics that can be related to a combination of a rotation and a parity transformation. Specifically, the swap of components $a_i$ and $a_j$ resembles a rotation in the $i$-$j$ plane, while the inversion of the signs resembles a parity transformation. 
This suggests that the specific higher-order perturbation is consistent with a system that exhibits rotational symmetry in the $i$-$j$ plane. It indicates that the $i$-th and $j$-th dimensions represent interchangeable semantic features, such that rotating the embedding vector in the $i$-$j$ plane does not significantly alter its semantic meaning.

\section{Quantum Mechanical Interpretation and Time Evolution}
\label{sec_quantum_interpretation}

In the previous section, we established a classical Hamiltonian representation of semantic transformations within LLM embedding spaces. We now explore how to extend this framework to incorporate concepts inspired by quantum mechanics, specifically by introducing unitary transformations. This allows us to draw analogies between the dynamics of semantic representations and the behavior of quantum systems.

\subsection{Unitary Transformation of the Hamiltonian}

Starting with the transformed cosine similarity $S_C' = \mathbf{a}^T \mathbf{H'} \mathbf{a}$, we seek to express this relationship in a form that allows further analysis. Since $\mathbf{H'}$ is a real, symmetric matrix, it can be diagonalized by an orthogonal (and therefore unitary) transformation. Let $\mathbf{U}$ be a unitary matrix such that

\begin{equation}
    \mathbf{D} = \mathbf{U}^\dagger \mathbf{H'} \mathbf{U}
\end{equation}

\noindent
where $\mathbf{D}$ is a diagonal matrix containing the eigenvalues of $\mathbf{H'}$. We can then define a new state vector $\ket{a'}$ as

\begin{equation}
    \ket{a'} = \mathbf{U}^\dagger \ket{a}
\end{equation}

\noindent
Substituting these expressions into the equation for $S_C'$, we obtain

\begin{equation}
    S_C' = \bra{a} \mathbf{H'} \ket{a} = \bra{a} \mathbf{U} \mathbf{D} \mathbf{U}^\dagger \ket{a} = \bra{a'} \mathbf{D} \ket{a'}
\end{equation}

\noindent
This transformation expresses the cosine similarity in terms of the eigenvalues of $\mathbf{H'}$ and the components of the transformed state vector $\ket{a'}$. The diagonal elements of $\mathbf{D}$ can be interpreted as representing distinct semantic features, while the components of $\ket{a'}$ represent the weighting of these features.

\subsection{Introducing Complex State Vectors and Time Evolution}

To explore potential dynamics within this framework, we now introduce complex coefficients and a time-dependent perspective. We replace the real components of $\ket{a'}$ with complex, time-dependent coefficients $c_n(t)$

\begin{equation}
    c_n(t) = A_n e^{-i E_n t / \hbar}
\end{equation}

\noindent
where
$A_n$ is a real amplitude, related to the initial value of the $n$-th component of $\ket{a'}$,
$E_n$ is the $n$-th eigenvalue of $\mathbf{H'}$ (a diagonal element of $\mathbf{D}$),
$t$ is time,
$\hbar$ is a scaling constant introduced for dimensional consistency, and
$i$ is the imaginary unit.

We then define a complex, time-dependent state vector $\ket{\psi(t)}$ as

\begin{equation}
    \ket{\psi(t)} = \sum_n c_n(t) \ket{n}
\end{equation}

\noindent
where $\ket{n}$ are the eigenvectors of $\mathbf{H'}$ (the columns of the matrix $\mathbf{U}$).
The expectation value associated with $\mathbf{H'}$ becomes:

\begin{equation}
    \langle H'(t) \rangle = \bra{\psi(t)} \mathbf{H'} \ket{\psi(t)} = \sum_n |c_n(t)|^2 E_n = \sum_n A_n^2 E_n
\end{equation}

\noindent
In this simplified model, the expectation value is constant in time because the amplitudes $A_n$ are constant. However, the introduction of complex phases allows for the possibility of more complex dynamics if we were to introduce a time-dependent Hamiltonian.

This framework can be related to the concept of a Time-Dependent Schrödinger Equation

\begin{equation}
    i \hbar \frac{d}{dt} \ket{\psi(t)} = \mathbf{H'} \ket{\psi(t)}
\end{equation}

\noindent
This can be verified by substituting the expression for $\ket{\psi(t)}$ and using the fact that $\mathbf{H'} \ket{n} = E_n \ket{n}$.
By introducing unitary transformations and complex coefficients, we have created a model of semantic transformations within LLM embedding spaces that allows us to explore potential dynamics. 

\subsection{Diagonalization of the Hamiltonian H'}

In this subsection, we aim to diagonalize the Hamiltonian matrix $\mathbf{H'}$ derived from the cosine similarity expression. Recall that $\mathbf{H'}$ is defined as

\begin{equation}
\mathbf{H'} = \frac{1}{\sum_{i=1}^N v_i^2}
\begin{bmatrix}
    v_1^2 & v_1 v_2 & \cdots & v_1 v_N \\
    v_2 v_1 & v_2^2 & \cdots & v_2 v_N \\
    \vdots & \vdots & \ddots & \vdots \\
    v_N v_1 & v_N v_2 & \cdots & v_N^2
\end{bmatrix}
\end{equation}

\subsubsection{Eigenvalues of $\mathbf{H'}$}

The matrix $\mathbf{H'}$ can be expressed more compactly as

\begin{equation}
\mathbf{H'} =  \frac{1}{\sum_{i=1}^N v_i^2} \mathbf{v} \mathbf{v}^T
\end{equation}

\noindent
where $\mathbf{v} = [v_1, v_2, ..., v_N]^T$ is a column vector. This form reveals the rank-1 nature of $\mathbf{H'}$.
A matrix of the form $\mathbf{H'} = k \mathbf{v} \mathbf{v}^T/||\mathbf{v}||^2$ (where $k$ is a scaling factor) has a specific eigenvalue structure. In our case, $k=1$.
The matrix $\mathbf{H'}$ has one eigenvalue equal to 1, and all other eigenvalues are 0.
This is because the trace of $\mathbf{H'}$ is

\begin{equation}
Tr(\mathbf{H'}) = \frac{1}{\sum_{i=1}^N v_i^2} \sum_{i=1}^N v_i^2 = 1
\end{equation}

\noindent
Since the trace is the sum of the eigenvalues, and we know the matrix is rank-1, that eigenvalue must be 1.
Therefore, the diagonalized form of $\mathbf{H'}$, denoted as $\mathbf{D}$, is a diagonal matrix with one entry equal to 1 and all other entries equal to 0

\begin{equation}
\mathbf{D} =
\begin{bmatrix}
    1 & 0 & \cdots & 0 \\
    0 & 0 & \cdots & 0 \\
    \vdots & \vdots & \ddots & \vdots \\
    0 & 0 & \cdots & 0
\end{bmatrix}
\end{equation}

\noindent
In summary, by recognizing the structure of $\mathbf{H'}$ as a rank-1 matrix, we can directly determine its eigenvalues and diagonalized form, simplifying further analysis.

\subsubsection{Eigenvectors of $\mathbf{H'}$}

The Hamiltonian matrix $\mathbf{H'}$ possesses a specific set of eigenvectors that correspond to its eigenvalues. For the eigenvalue $\lambda_1 = 1$, the normalized eigenvector is given by

\begin{equation}
    \mathbf{x}_1 = \frac{\mathbf{v}}{||\mathbf{v}||} = \frac{\mathbf{v}}{\sqrt{\sum_{i=1}^N v_i^2}}
\end{equation}

\noindent
This eigenvector is simply the vector $\mathbf{v}$ normalized to unit length.
For the eigenvalue $\lambda_i = 0$ (where $i$ ranges from 2 to $N$), the corresponding eigenvectors $\mathbf{x}_i$ must satisfy the condition

\begin{equation}
    \mathbf{v}^T \mathbf{x}_i = 0
\end{equation}

\noindent
This condition implies that these eigenvectors are orthogonal to the vector $\mathbf{v}$. Crucially, these eigenvectors must also be linearly independent and mutually orthogonal to each other to form a complete basis. The Gram-Schmidt process can be used to construct such a set of orthogonal eigenvectors.

\subsubsection{Unitary Matrix $\mathbf{U}$}

The unitary matrix $\mathbf{U}$ that diagonalizes $\mathbf{H'}$ is constructed by using the normalized eigenvectors as its columns

\begin{equation}
    \mathbf{U} = [\mathbf{x}_1, \mathbf{x}_2, ..., \mathbf{x}_N]
\end{equation}

\noindent
Here, $\mathbf{x}_1$ is the normalized eigenvector corresponding to the eigenvalue 1, and $\mathbf{x}_2$ through $\mathbf{x}_N$ are the $N-1$ eigenvectors corresponding to the eigenvalue 0. The orthogonality of these eigenvectors ensures that $\mathbf{U}$ is a unitary matrix, which preserves the norm of vectors under transformation.

\subsection{Time Evolution with Diagonalized Hamiltonian}

Having obtained the diagonalized Hamiltonian $\mathbf{D}$, we can now analyze the time evolution of the system.
Let's consider the case where the eigenvalues are $E_1 = 1$ and $E_n = 0$ for $n = 2, 3, ..., N$. This significantly simplifies the expressions.
The time-dependent coefficients become

\begin{align}
    c_1(t) &= A_1 e^{-i (1) t / \hbar} = A_1 e^{-i t / \hbar} \\
    c_n(t) &= A_n e^{-i (0) t / \hbar} = A_n \quad \text{for } n = 2, 3, ..., N
\end{align}

\noindent
This means that only the first coefficient, $c_1(t)$, acquires a time-dependent phase. All other coefficients remain constant in time.
The state vector becomes

\begin{equation}
    \ket{\psi(t)} = A_1 e^{-i t / \hbar} \ket{1} + \sum_{n=2}^N A_n \ket{n}
\end{equation}

\noindent
The expectation value simplifies to

\begin{equation}
    \langle H'(t) \rangle = |c_1(t)|^2 (1) + \sum_{n=2}^N |c_n(t)|^2 (0) = A_1^2
\end{equation}

\noindent
Since $A_1$ is constant, the expectation value is also constant in time.
The fact that we have one eigenvalue equal to 1 and $N-1$ eigenvalues equal to 0 has several important consequences:

\begin{enumerate}
    \item Single Active Mode: The system effectively has only one mode that evolves in time. This mode is associated with the eigenvector $\ket{1}$ corresponding to the eigenvalue 1. All other modes remain constant.
    \item Simplified Dynamics: The dynamics of the system are greatly simplified. The time evolution is governed by a single phase factor, $e^{-i t / \hbar}$, associated with the active mode.
    \item Constant Value: The expectation value is constant and equal to the square of the amplitude of the active mode.
    \item Projection onto a Subspace: The system's state vector can be seen as a projection onto a one-dimensional subspace spanned by the eigenvector $\ket{1}$, plus a constant component in the orthogonal subspace. The time evolution only affects the component in the subspace spanned by $\ket{1}$.
\end{enumerate}

\noindent
In the context of LLM embedding spaces, this eigenvalue structure suggests that the semantic transformation is dominated by a single, time-evolving feature, while other features remain relatively static. This could correspond to a scenario where a particular aspect of the semantic meaning is changing over time, while other aspects remain constant.

\subsection{Analogue of Zero-Point Energy}

Within our defined LLM Embedding System, the cosine similarity $S_C$ is strictly bounded between -1 and 1, and consequently, the transformed cosine similarity $S_C' = \frac{1}{2}(S_C + 1)$ is strictly bounded between 0 and 1. Specifically, $0 < S_C' < 1$. 
This was the assumption of the LLM Embedding Model.
Therefore, the minimum possible value of $S_C'$ is not zero, but rather a small positive value, which we denote as $\epsilon > 0$. This value represents the closest possible approach to maximal dissimilarity within our system

\begin{equation}
S_{C,min}' = \epsilon > 0
\end{equation}

\noindent
This non-zero minimum value can be interpreted as an analogue of zero-point energy. It reflects the inherent semantic content that is present even in the state of near-maximal dissimilarity. Even the most dissimilar concepts within our defined system still share some underlying semantic features, preventing them from being completely orthogonal in the embedding space.

To quantify this analogue, we can express the transformed cosine similarity in terms of the Hamiltonian $\mathbf{H'}$ and the state vector $\ket{a}$

\begin{equation}
S_C' = \bra{a} \mathbf{H'} \ket{a} = \epsilon
\end{equation}

\noindent
This equation implies that the expectation value associated with $\mathbf{H'}$ in the state $\ket{a}$ is equal to $\epsilon$.
To further refine our understanding, let's decompose the state vector $\ket{a}$ into two components: one component that is aligned with the eigenvector $\ket{1}$ corresponding to the non-zero eigenvalue of $\mathbf{H'}$, and another component that is orthogonal to this eigenvector

\begin{equation}
\ket{a} = A_1 \ket{1} + \ket{\delta}
\end{equation}

\noindent
where $A_1$ is the amplitude of the component aligned with $\ket{1}$, and $\ket{\delta}$ is the component orthogonal to $\ket{1}$.
The analogue of zero-point energy can then be expressed as

\begin{equation}
E_{ZP} = \bra{a} \mathbf{H'} \ket{a} = \bra{A_1 \ket{1} + \ket{\delta}} \mathbf{H'} \bra{A_1 \ket{1} + \ket{\delta}} = A_1^2 + \bra{\delta} \mathbf{H'} \ket{\delta}
\end{equation}

\noindent
Since $\ket{\delta}$ is orthogonal to $\ket{1}$, we have $\bra{\delta} \mathbf{H'} \ket{\delta} = 0$. Therefore, the analogue simplifies to

\begin{equation}
E_{ZP} = A_1^2 = \epsilon
\end{equation}

\begin{figure}[htpb]
    \centering
    \includegraphics[width=0.65\textwidth]{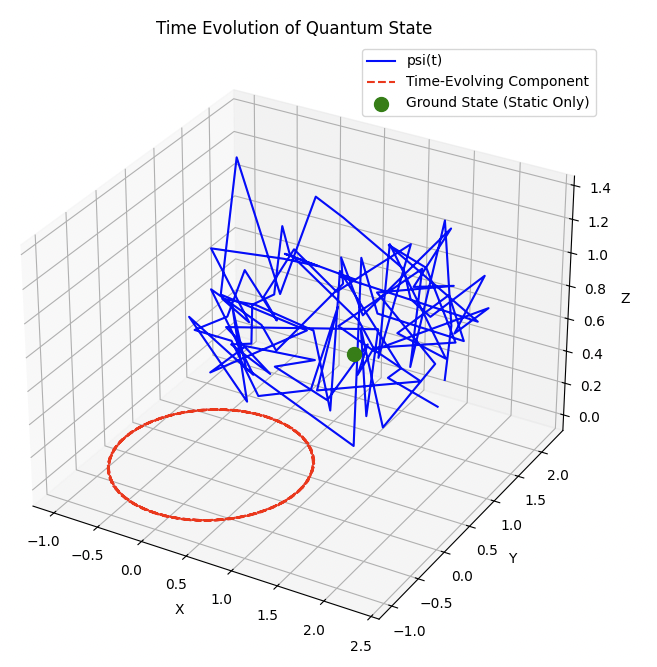}
\caption{Visualization of the time evolution of the quantum state $\ket{\psi(t)}$. The x and y axes represent the real and imaginary components of the time-evolving term $A_1 e^{-i t / \hbar} \ket{1}$, tracing a circle in the complex plane (red dashed line). The z-axis represents a simplified sum of static components $\sum_{n=2}^N A_n \ket{n}$. The blue line shows the trajectory of the superposition $\ket{\psi(t)}$. The green sphere indicates the ground state, consisting only of static components.}
    \label{fig_quantum_state}
\end{figure}

\noindent
This result implies that the analogue is directly proportional to the square of the amplitude of the component of the state vector that is aligned with the eigenvector corresponding to the non-zero eigenvalue of $\mathbf{H'}$. In other words, the analogue is determined by the degree to which the state vector is aligned with the dominant mode of the system. An illustrative example of a time-dependent quantum state with ground state energy is shown in Figure~\ref{fig_quantum_state}.

The amplitude $A_1$ can be derived from the transformed cosine similarity $S_{CN}'$ as follows, Eq.~(\ref{eq_cossimall3})

\begin{equation}
    A_1 = \sqrt{\epsilon} = \sqrt{\min\left( \frac{1}{\sum_{i=1}^N v_i^2}\left (\sum_{i=1}^N v_ia_i\right )^2 \right)}
\end{equation}

\noindent
where $\epsilon$ is the minimum possible value of $S_{CN}'$, representing the analogue of zero-point energy. 

In the context of LLM embedding spaces, this analogue could be interpreted as the minimum level of semantic coherence that is required for a concept to be considered a physical embedding within our defined system. It represents the inherent semantic content that prevents a concept from being completely meaningless or random. The value of $\epsilon$ would depend on the specific LLM architecture and training data, reflecting the inherent structure of the embedding space.

\subsection{Analogy to the Quantum Harmonic Oscillator and Koopman-von Neumann Mechanics}

Having derived a framework for LLM embedding spaces, it is instructive to draw parallels to other well-established systems. The transformation we have performed, starting from an LLM Embedding System and arriving at a representation using linear algebra, shares similarities with the quantum harmonic oscillator and the Koopman-von Neumann (KvN) mechanics.

\subsubsection{Quantum Harmonic Oscillator}

The quantum harmonic oscillator is a fundamental system in quantum mechanics that describes a particle subject to a restoring force proportional to its displacement from equilibrium. Key features of this system include:

\begin{enumerate}
    \item Discrete Energy Levels: The energy of the quantum harmonic oscillator is quantized, meaning it can only take on discrete values.
    \item Zero-Point Energy: Even in its ground state, the quantum harmonic oscillator possesses a non-zero energy, known as the zero-point energy.
    \item Wave-like Behavior: The quantum harmonic oscillator is described by a wave function that evolves in time according to the Schrödinger equation.
\end{enumerate}

\noindent
In our LLM Embedding System, the eigenvalues of the Hamiltonian $\mathbf{H'}$ can be interpreted as representing distinct semantic features, analogous to the discrete energy levels of the quantum harmonic oscillator. The L2 normalization constraint and the structure of the embedding space impose a form of organization that leads to these distinct features. The single non-zero eigenvalue suggests a dominant semantic mode, similar to the ground state of the harmonic oscillator. Furthermore, we have derived an analogue to zero-point energy, $E_{ZP} = \epsilon$, representing the minimum semantic content present even in states of near-maximal dissimilarity. The complex coefficients in our time-dependent model can be viewed as representing oscillations between different semantic states, analogous to the oscillations of the quantum harmonic oscillator.

\subsubsection{Koopman-von Neumann Mechanics}

Koopman-von Neumann (KvN) mechanics is an alternative formulation of classical mechanics that uses mathematical tools similar to those used in quantum mechanics. In KvN mechanics, classical states are represented as vectors in a Hilbert space, and classical observables are represented as operators acting on these vectors.
The key difference between KvN mechanics and standard classical mechanics is that KvN mechanics linearizes the classical dynamics by embedding the classical phase space into a larger Hilbert space. This linearization allows for the application of linear algebra techniques to analyze the classical system.
Our approach to LLM embedding spaces shares several similarities with KvN mechanics:

\begin{enumerate}
    \item Linearization: We have effectively linearized the complex relationships within the LLM embedding space by representing semantic transformations as linear operations in a high-dimensional vector space.
    \item Hilbert Space: The L2-normalized embedding vectors reside in a Hilbert space, allowing us to apply linear algebra tools.
    \item Transformations: The use of transformations to diagonalize the Hamiltonian is related to the unitary time evolution operator in KvN mechanics. The single dominant eigenvalue in our diagonalized Hamiltonian suggests a simplified KvN system where the dynamics are largely governed by a single mode.
\end{enumerate}

\noindent
However, it is important to note that our approach is not a direct application of KvN mechanics. We are not starting with a classical system and then linearizing it using the KvN formalism. Instead, we are starting with a system (LLM embedding spaces) that is already represented as vectors in a high-dimensional space and then applying linear algebra tools to analyze its structure and dynamics. The KvN analogy provides a useful framework for understanding how linear algebra can be used to analyze complex systems.

\section{Potential Applications: Mitigating Hallucinations}

One potential application of our framework lies in mitigating hallucinations in LLMs. We hypothesize that hallucinations arise from transitions to semantically incoherent states within the embedding space. These incoherent states may be characterized by:

\begin{enumerate}
    \item Distorted Eigenvector Distribution: Transitions to states where the dominant eigenvalue is 1, but the distribution of the corresponding eigenvector's components is significantly different from that of coherent semantic representations. This could indicate an over-reliance on a single, potentially irrelevant, feature or a distorted combination of features.
    \item Unstable Combinations: Combinations that are highly sensitive to small perturbations, leading to unpredictable outputs.
    \item Violation of Structure: Hallucinations might involve semantic representations that violate the parity or rotational symmetries observed in coherent states.
\end{enumerate}

Based on these characteristics, we propose the following strategies for mitigating hallucinations:

\begin{enumerate}
    \item Eigenvector Distribution Regularization: Introduce a regularization term in the LLM training objective that penalizes transitions to states with distorted eigenvector distributions. This could involve minimizing the difference between the eigenvector distribution of the current state and the average eigenvector distribution of coherent semantic representations. This would encourage the model to favor combinations of features that are more typical of meaningful content.
    \item Symmetry Enforcement: Develop techniques to explicitly enforce parity and rotational symmetries during LLM training and inference. This could involve adding constraints to the embedding vectors or modifying the attention mechanisms to respect these symmetries.
    \item Uncertainty Quantification: Adapt techniques for uncertainty quantification to estimate the likelihood of a hallucination. This could involve calculating the variance.
\end{enumerate}

\noindent
Further research is needed to validate these strategies and to develop practical algorithms for implementing them. However, the framework provides a new perspective on the problem of hallucinations and suggests potentially fruitful avenues for future investigation.

\section{Discussions}

This article has presented a new approach to analyzing LLM embedding spaces, drawing inspiration from mathematical tools. We have demonstrated that the L2 normalization constraint, a common characteristic of LLM architectures, leads to a structured embedding space with properties that can be effectively explored using mathematical formalisms, particularly the Hamiltonian formalism. Our analysis has revealed relationships between LLM embedding spaces and other systems, suggesting that these tools may provide valuable insights into the inner workings of LLMs.

One potential application of this framework lies in addressing the challenge of hallucinations, where LLMs generate factually incorrect or nonsensical outputs. If the mathematical structures of LLM embedding spaces exhibit similarities, then insights might offer new approaches to managing this uncertainty.

The relationships between LLM embedding spaces and other systems could facilitate the use of advanced computing to accelerate LLM training and inference. Algorithms have the potential to provide performance gains for tasks such as linear algebra and optimization, which are central to LLM operation.

This work offers a promising avenue for gaining deeper insights into LLMs. The relationships we have drawn, while requiring careful consideration, provide a perspective on the structure and behavior of LLM embedding spaces. The Hamiltonian formalism provides a valuable tool for analyzing semantic relationships and for understanding how LLMs process and represent complex semantic information.

It is important to emphasize that we do not assert that specific symmetries are necessarily present in LLM Embedding Systems. However, our analysis reveals no contradictions that would preclude their existence, suggesting a potential avenue for future research.

While the framework presented here is still in its early stages of development, it offers a compelling vision for the future of LLM research. Exploration of these connections could lead to transformative advancements in both the efficiency and reliability of LLMs. The work should focus on developing more nuanced similarity measures, investigating non-linear effects, analyzing the Hamiltonian spectrum, and exploring the relationship between attention weights and Hamiltonian elements. Ultimately, the goal is to develop algorithms and techniques that can address the key challenges within LLMs, such as computational cost and reliability.

\end{document}